\documentclass[runningheads]{llncs}
\usepackage{xcolor}
\usepackage{booktabs}
\usepackage{graphicx}
\usepackage{hyperref}
\usepackage{graphicx}
\usepackage{multirow} 
\usepackage{fontawesome5}
\usepackage{amsmath,amssymb}

\usepackage[ruled,linesnumbered]{algorithm2e}

\begin{document}
\title{Advancing Aspect-Based Sentiment Analysis through Deep Learning Models}
\author{Chen Li \inst{1} 
\and Huidong Tang \inst{2} 
\and Jinli Zhang \inst{3}\faIcon{envelope}
\and Xiujing Guo \inst{4} 
\and Debo Cheng \inst{5} 
\and Yasuhiko Morimoto \inst{6}
}
\authorrunning{C. Li et al.}
\institute{
Graduate School of Informatics, Nagoya University, Japan \\
Shandong Xiehe University, China \\
Beijing University of Technology, Beijing, China \\
\faIcon{envelope} \email{jlzhangcs@bjut.edu.cn} \\
Graduate School of Information Science and Technology, Osaka University, Japan \\
UniSA STEM, University of South Australia, Australia \\
Graduate School of Advanced Science and Engineering, Hiroshima University, Japan
}
\maketitle 

\begin{abstract}
Aspect-based sentiment analysis predicts sentiment polarity with fine granularity. While graph convolutional networks (GCNs) are widely utilized for sentimental feature extraction, their naive application for syntactic feature extraction can compromise information preservation. This study introduces an innovative edge-enhanced GCN, named SentiSys, to navigate the syntactic graph while preserving intact feature information, leading to enhanced performance. Specifically,we first integrate a bidirectional long short-term memory (Bi-LSTM) network and a self-attention-based transformer. This combination facilitates effective text encoding, preventing the loss of information and predicting long dependency text. A bidirectional GCN (Bi-GCN) with message passing is then employed to encode relationships between entities. Additionally, unnecessary information is filtered out using an aspect-specific masking technique. To validate the effectiveness of our proposed model, we conduct extensive evaluation experiments on four benchmark datasets. The experimental results demonstrate enhanced performance in aspect-based sentiment analysis with the use of SentiSys.

\keywords{Aspect-based sentiment analysis systems \and Graph convolutional networks}
\end{abstract}

\section{Introduction}
\label{sec:introduction}
Aspect-based sentiment analysis \cite{laskari2016aspect} aims to classify sentences in aspect granularity. The sentiments of each aspect can be different even in the same sentence. Aspect-level sentiment analysis has been widely used in real-world applications. Suppose there is such a review comment among the restaurant reviews: “Although the menu is limited, the friendly staff provided us with a nice night." If we pay attention to the aspect of “menu", the sentiment is negative (“limited"); when it comes to “staff", we can extract the opposite sentiment, i.e., positive (“friendly"). Such multi-aspect can introduce multi-sentiments in the same sentence, resulting in difficulty of sentiment prediction. To solve such difficulty, aspect-based sentiment analysis is attracting attention \cite{bauman2022know}.

Neural networks are widely used for aspect-based sentiment analysis because of their promising results, especially recurrent neural networks (RNNs) \cite{zhang2019multi} and the Transformer \cite{li2024tengan,vaswani2017attention} have demonstrated their sequence modeling capabilities. However, these models are not affordable to establish precise dependencies between word pairs and aspects. Suppose such a sentence: “The dish looks mediocre but tastes surprisingly wonderful!" Although the aspect “dish" is linked to two opposite sentiments, “looks mediocre" and “tastes surprisingly wonderful", we can easily understand that the latter is more tightly linked to the “dish". However, this judgment may be difficult for these models due to the closer distance of the former to the “dish". Convolutional neural networks (CNNs) \cite{rani2022efficient,zhang2024cd} are studied to make predictions based on phrases that express sentiments more accurately. However, phrases generally have longer dependencies on the subject. For example, “tastes surprisingly wonderful" may express a positive sentiment more accurately than just “wonderful". CNNs are not affordable for such long dependencies because of the local receptive field limitation \cite{kipf2016semi}.

Graph convolutional networks (GCNs) \cite{zhang2019predicting,zhou2020graph} can decompose sentences into a tree structure composed of entities. Such a tree structure constructs a more direct path between aspects and related words. Therefore, GCNs resolve the long-distance dependency problem and facilitate the learning of the representation of nodes and their positional relationships \cite{meng2020structure}. However, the general GCNs lack the edge information. GCNs cannot capture the syntax and semantic relationships due to the lack of edge information resulting in poor performance \cite{chang2023reducing}. This study introduces an innovative edge-enhanced graph convolutional network (GCN) named SentiSys, designed to elevate the effectiveness of aspect-based sentiment analysis systems. In our approach, we leverage bidirectional long short-term memory (Bi-LSTM) to extract node word embeddings, and Bi-GCN is employed to model word dependencies within a tree structure, thereby enhancing word representations. The integration of these components facilitates accurate association of sentiment words with the target aspect, resulting in a notable performance improvement. The robustness of the proposed SentiSys is substantiated through extensive evaluation, confirming its high-performance capabilities. The main contributions are as follows:
\begin{itemize}
\item {\bf Architecture based on neural networks:} SentiSys integrates Bi-LSTM, a transformer, and an edge-enhanced Bi-GCN, leveraging linguistic syntax to construct a graph. This graph serves as the foundation for extracting and representing textual content through a sophisticated GCN. This approach highlights the significance of harnessing these advanced technologies to decode sentiment and meaning within text, marking a notable advancement in sentiment analysis.
\item {\bf Effective semantic edges:} Using a dependency tree minimizes the gap between aspects and related words while establishing a syntactic pathway within the text. SentiSys also considers edge information within syntactic dependency trees, enhancing syntactic connections between associated words and the target aspect.
\item {\bf Performance enhancement:} The proposed methodology's efficacy is confirmed through validation on four standard benchmark datasets. The empirical findings conclusively illustrate that the SentiSys model surpasses other baseline models across the majority of performance metrics.
\end{itemize}

\section{Related Work}
\label{sec:related}
\subsection{Aspect-based Sentiment Analysis Systems}
Conventional sentiment analysis methodologies typically involve training classifiers utilizing features such as bag-of-words and sentiment dictionaries. However, both rule-based \cite{ding2008holistic} and statistical-based strategies \cite{jiang2011target} require handcrafted features, revealing the limitations of these naive sentiment analysis methods, particularly when confronted with the complex intricacies inherent in customer review analysis. These challenges arise in capturing nuanced sentiments regarding specific aspects.

Motivated by the inadequacies of traditional sentiment analysis, the research landscape has progressively shifted towards aspect-based sentiment analysis. Aspect-based sentiment analysis encompasses four key tasks \cite{cai2020aspect}: aspect term sentiment analysis (ATSA), aspect category sentiment analysis (ACSA), aspect term extraction (ATE), and aspect category extraction (ACE). ATSA involves training classifiers to predict the sentiment polarity associated with specified aspect terms, addressing the need for a more granular understanding of sentiment expressions within the context of specific elements mentioned in reviews \cite{phan2020modelling}. ACSA, on the other hand, aims to predict sentiment polarity concerning predefined sentiment categories, providing insights into broader sentiment trends related to predetermined aspects \cite{liao2021improved}. ATE and ACE, conceptualized as entity extraction challenges, strive to solve the intricate problem of identifying and understanding aspects within reviews. ATE focuses on extracting precise aspect terms, such as the previously mentioned example of “dish," aiming to capture specific entities of interest \cite{augustyniak2021comprehensive}. Conversely, ACE aims to identify and categorize broader aspects, like the “food" category, encompassing terms like “dish."

In addressing these tasks, the motivation lies in achieving a more nuanced and targeted comprehension of sentiment expressions within diverse contexts. By refining the understanding of sentiments associated with specific aspects, aspect-based sentiment analysis contributes to enhanced accuracy and applicability in sentiment analysis practices \cite{do2019deep}. Recognizing limitations is crucial, particularly in developing robust methods to address ambiguous context, thereby prompting further research to enhance the efficacy of aspect-based sentiment analysis.

\subsection{RNN-based Sentiment Analysis Systems}
The recent remarkable advances in neural networks have attracted the attention of researchers for applications \cite{li2023spotgan,li2024gxvaes,zhang2024quantitative}. The models proposed in \cite{tang2016effective} for target-dependent sentiment classification leverage neural networks. Specifically, the LSTM architecture, originally designed for sequence modeling, proves effective for aspect-based sentiment analysis tasks in an autoregressive manner. Building upon the LSTM framework, this approach enhances the architecture by incorporating target-specific information. Referred to as target-dependent LSTM (TD-LSTM), this model skillfully captures the dynamic interplay between a target word and its contextual words, discerningly selecting pertinent segments of the context to deduce sentiment polarity towards the target. The model is amenable to end-to-end training through standard backpropagation, employing the cross-entropy error of supervised sentiment classification as its loss function. While the TD-LSTM approach enhances sentiment classification by considering target-specific information, the nuanced understanding of the interplay between a target word and its contextual words remains a complex task. This limitation affects model accuracy in capturing subtle sentiment nuances and context-specific variations.

Similar to TD-LSTM, ATAE-LSTM \cite{wang2016attention} is crafted to emphasize contextual relationships, utilizing a Bi-LSTM and an attention-based LSTM, respectively. While these models exhibit strengths, they encounter difficulties in capturing extensive contextual information, especially when crucial details are distant from the target. Recognizing this limitation, RAM \cite{chen2017recurrent} and TNet-LF \cite{li2018transformation} extend ATAE-LSTM \cite{wang2016attention} by introducing a multi-attention and Bi-attention mechanism, respectively. These enhancements significantly augment the models' capability to consider broader contextual information, effectively addressing the challenge of handling long-range dependencies. However, TNet-LF cannot fully capture the intricacies of sentence structure and grammar, limiting its ability to discern nuanced syntactic relationships crucial for a comprehensive understanding of sentiment in text. Additionally, its reliance on semantic aspects may result in suboptimal performance when dealing with complex syntactic structures in natural language. Sentic LSTM \cite{ma2018targeted} stands as a sophisticated system designed for targeted aspect-based sentiment analysis. The core innovation lies in the infusion of commonsense knowledge into an attentive LSTM network. RACL \cite{chen2020relation} is a sentiment analysis system based on relation-aware collaborative learning. This system tackles fundamental sentiment analysis tasks, encompassing aspect term extraction, opinion term extraction, and aspect-level sentiment classification. Nevertheless, the absence of detailed assessments and comparisons with existing models impedes a comprehensive understanding of RACL's effectiveness in real-world scenarios.

\subsection{GCN-based Sentiment Analysis Systems}
Unlike RNN-based sentiment analysis systems, GCN-based systems seek to represent language data as graphs, where words or phrases are nodes and relationships between them are edges. This representation enables a nuanced understanding of contextual dependencies, capturing intricate connections challenging for traditional RNNs \cite{wang2020relational}.

Several notable advancements in aspect-based sentiment analysis using GCNs have been introduced. ASGCN \cite{zhang2019aspect} showcases remarkable proficiency in capturing sentiment polarity across various aspects in a text. This is achieved through its adept modeling of relationships between aspects and their respective words utilizing GCNs. However, the efficacy of ASGCN is contingent on precise dependency tree parsing. Instances of inaccurate parsing pose a potential challenge, introducing a risk of suboptimal modeling. This directly undermines the fidelity of relationships between aspects and words. Consequently, the overall performance of ASGCN is adversely affected, especially within the domain of aspect-based sentiment analysis. 

Unlike prior research, we introduce a novel fusion of Bi-LSTM, Transformer, and GCNs to advance aspect-based sentiment analysis. SentiSys strategically utilizes each component's strengths: Bi-LSTM acts as an adept feature extractor, capturing intricate patterns and semantic nuances. Meanwhile, the Transformer excels at capturing long-range dependencies, enabling our model to understand crucial contextual relationships. The introduction of GCN with a dependency parsing layer enriches the model by incorporating syntactic information, facilitating a more nuanced understanding of text syntax. By combining these elements, SentiSys aims to offer a comprehensive perspective on aspect-based sentiment analysis, addressing previous limitations and fostering a more accurate sentiment analysis methodology.

\section{SentiSys}
\label{sec:model}

\begin{figure}[t]  
\centering  
\includegraphics[width=0.85\textwidth]{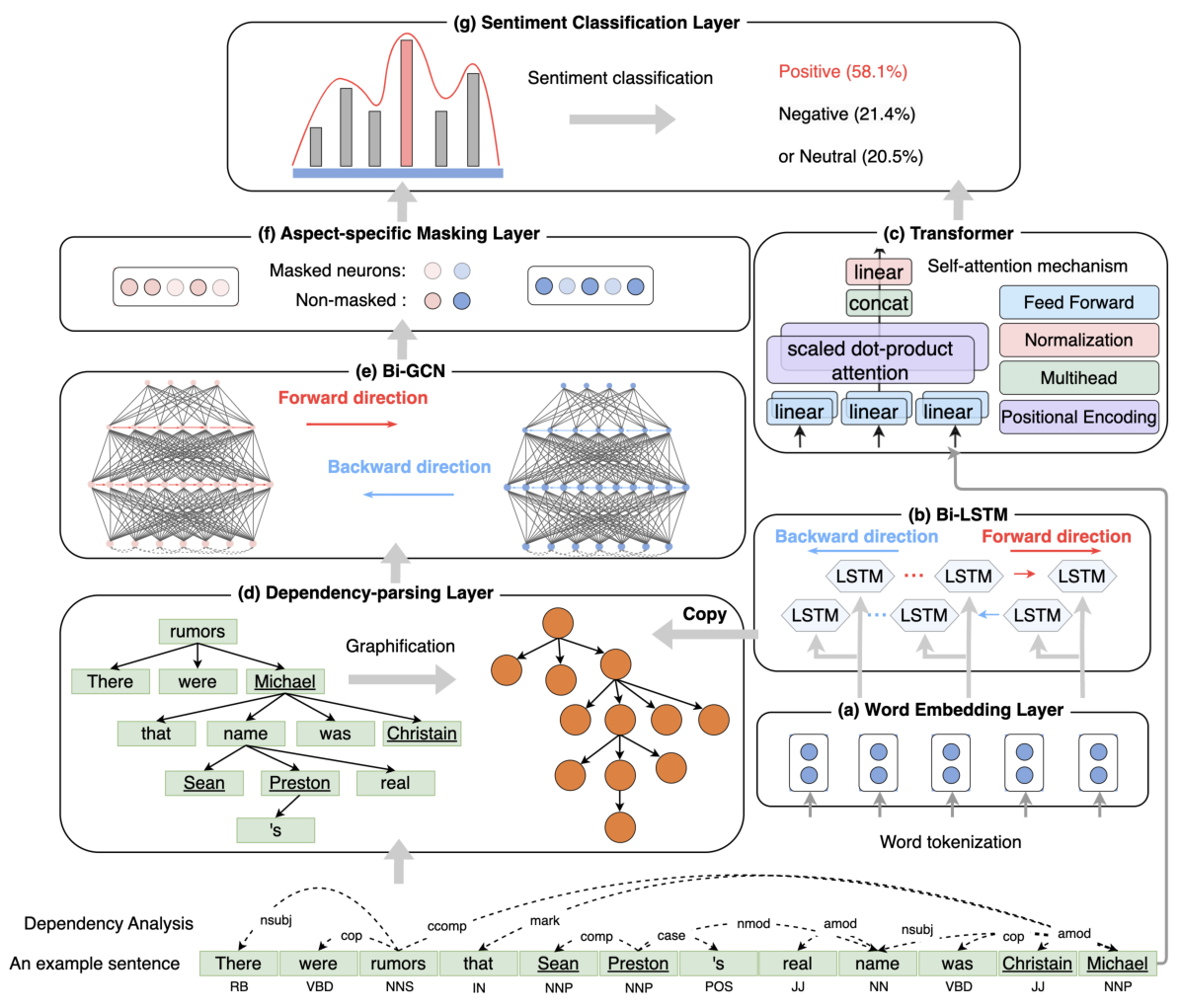}   
\caption{SentiSys for aspect-based sentiment analysis includes the following components: (a) Initial word embedding layer for parsing and embedding words into vectors. (b) Bi-LSTM network captures contextual information for deeper word relationship understanding. (c) Transformer layer with self-attention mechanism explores global word features, enhancing contextual understanding. (d) Dependency parsing layer recognizes word dependencies, constructing a dependency tree to explain grammatical structure. (e) Bi-GCN propagates information among tree nodes, effectively modeling word relationships. (f) Aspect-specific masking refines hidden representation, reducing redundancy and improving accuracy. (g) Sentiment classification layer categorizes sentence sentiment into predefined categories, such as positive, negative, or neutral.}
\label{fig:overview} 
\end{figure}

SentiSys consists of a word embedding layer for token representation, a Bi-LSTM for sentence feature extraction, a Transformer for global contextual feature extraction, a dependency parsing layer for context grammar tree creation, a Bi-GCN for processing the tree structure, an aspect-specific masking layer to reduce redundancy and enhance accuracy, and a sentiment classification layer for sentence sentiment classification. Figure \ref{fig:overview} illustrates the architecture of SentiSys for aspect-based sentiment analysis.

\subsection{Word Embedding Layer}
\label{sec:embedding}
To tokenize a sentence consisting of $n$ words with an $m$-word aspect, we represent it as a sequence denoted as 
${\bf S} = \left\{{\bf w}_1, {\bf w}_2, \cdot\cdot\cdot, {\bf w}_{\gamma}, \cdot\cdot\cdot, {\bf w}_{\gamma+m}, \cdot\cdot\cdot, {\bf w}_n \right\}$, where ${\bf w}_t \in {\bf S}$ and $\gamma$ indicate a word token and starting index of an aspect, respectively. Each word token is then transformed into a low-dimensional vector denoted as ${\bf e}_t \in \mathbb{R}^{d_w}$, where $d_w$ denotes the dimension of the word embedding.

\subsection{Bi-LSTM Network}
\label{sec:bi-lstm}
Word embedding embeds the words in the sentence into low-dimensional vectors, and these embedding vectors are used as inputs to the Bi-LSTM network \cite{graves2013speech}. The utilization of Bi-LSTM allows for the extraction of contextual information from sentences. As mentioned in \cite{hu2021graph}, the Bi-LSTM network comprises two components: the forward LSTM and the backward LSTM. In the forward LSTM, the hidden vector $\overrightarrow {\bf h}^{LSTM}_t$ at the current time step is computed by combining the input embedding ${\bf e}_t$ with $\overrightarrow {\bf h}^{LSTM}_{t-1}$ from the previous time step. The backward LSTM is constructed by utilizing ${\bf e}_t$ and $\overleftarrow {\bf h}^{LSTM}_{t+1}$ from the subsequent time step. Finally, the hidden vectors $\overrightarrow {\bf h}^{LSTM}_t$ and $\overleftarrow {\bf h}^{LSTM}_t$ are concatenated to create the overall hidden vector ${\bf h}^{LSTM}_t$.
\begin{align}
\label{eq:lstm1}
\overrightarrow {\bf h}^{LSTM}_t = \text{LSTM}({\bf e}_t, \overrightarrow {\bf h}^{LSTM}_{t-1}),~\overrightarrow {\bf h}^{LSTM}_t \in \mathbb{R}^{d_h},\\
\setlength{\abovedisplayskip}{0pt}
\label{eq:lstm2}
\overleftarrow {\bf h}^{LSTM}_t = \text{LSTM}({\bf e}_t, \overleftarrow {\bf h}^{LSTM}_{t+1}),~\overrightarrow {\bf h}^{LSTM}_t \in \mathbb{R}^{d_h},\\
\setlength{\abovedisplayskip}{0pt}
\label{eq:lstm3}
{\bf h}^{LSTM}_t = \overrightarrow {\bf h}^{LSTM}_t \oplus \overleftarrow {\bf h}^{LSTM}_t,~{\bf h}^{LSTM}_t \in \mathbb{R}^{2 \times d_h},
\end{align}
where $d_h$ represents the dimension of the hidden layer in the Bi-LSTM, and $\oplus$ is the operator used for concatenation.

\subsection{Transformer Network}
\label{sec:transformer}
A transformer encoder, as described in \cite{vaswani2017attention}, utilizes self-attention mechanisms to examine the relationships and global characteristics of words within lengthy sentences. To incorporate the sequential positions of a sentence, sinusoidal positional encoding functions are employed, as outlined below:
\begin{align}
\label{eq:trans1}
\text{P}_{pos, 2i} = \sin (\frac{pos}{10000^{\frac{2i}{d_{model}}}}),\\
\setlength{\abovedisplayskip}{0pt}
\text{P}_{pos, 2i+1} = \cos (\frac{pos}{10000^{\frac{2i}{d_{model}}}}),
\end{align}
where $pos$ represents the position of a word within a sentence, while $i$ signifies the $i$th dimension of the embedding, with the embedding dimension denoted as $d_{model}$. Subsequently, the combined vector of the embedding and positional encoding for the input sentence serves as the input to the transformer. The transformer computes attention weights according to the following procedure:
\begin{align}
\label{eq:trans3}
&\text{Attention}({\bf Q, K, V}) = \text{Softmax}(\frac{{\bf QK}^T}{\sqrt{d_k}}){\bf V},\\
\setlength{\abovedisplayskip}{0pt}
\label{eq:trans4}
&\text{Multi-Head}({\bf Q, K, V}) = [\text{Head}_1, \cdots, \text{Head}_h] {\bf W},\\
\setlength{\abovedisplayskip}{0pt}
\label{eq:trans5}
&\text{Head}_i = \text{Attention}({\bf QW}_Q, {\bf KW}_K, {\bf VW}_V),
\end{align}
where ${\bf Q}$, ${\bf K}$ and ${\bf V}$ represent the query, key and value matrices with dimensions $d_k$, $d_k$ and $d_v$. ${\bf W, W}_Q, {\bf W}_K$, and ${\bf W}_V$ indicate weight matrices of the self-attention mechanism. Next, the transformer encoder applies two normalization layers to extract deep features of the context. Consider ${\bf Z}_{out}$ as the result matrix generated by the transformer encoder, and its formulation can be expressed as follows:
\begin{align}
\label{eq:trans6}
{\bf Z}_{out} = \text{Transformer}({\bf S}).
\end{align}
Finally, the calculated global attention scores play a pivotal role in the ultimate sentiment classification layer.
\subsection{Dependency-parsing Layer}
\label{sec:parsing}
The output hidden vector of the Bi-LSTM network, denoted as ${\bf h}_t$, can be used as an input to the dependency-parsing layer in the Bi-GCN, which helps to identify word dependencies in a sentence by constructing a dependency tree. Formally, the Bi-GCN initiates a matrix to record the edge information weights. Initially, it derives the adjacency matrix ${\bf A} \in \mathbb{R}^{n \times n}$ from the sentence's dependency tree. We present an adjacency matrix that contains non-discrete edge information, formulated as
\begin{equation}
\label{eq:parse2}
{\bf A} _{ij} =
\begin{cases}
1, \quad i=j, \\
\text{SDI}(i,j), \quad i \ne j,~i~\text{and}~j~\text{have dependencies}, \\
0, \quad \text{otherwise}.
\end{cases}
\end{equation}
Here, SDI is used to compute syntactic dependency information between the word $i$ and the word $j$. The computation is calculated as $\text{SDI}(i, j)= \frac{\text{Count}(\text{sd}(i, j))}{\text{Count}(\text{sd}(\cdot))}$, where $\text{sd}(i, j)$ and $\text{sd}(\cdot)$ represent the syntactic relationship between word pair $i$ and $j$ and the overall syntax of the dataset.

\subsection{Bi-GCN Network}
\label{sec:bi-gcn}
A GCN is a specialized variant of a Convolutional Neural Network (CNN) specifically tailored for encoding structured graph data. In the scenario of a textual graph consisting of $n$ words, we generate an adjacency matrix based on the syntactic matrix $\mathbf{A} \in \mathbb{R}^{n\times n}$ to capture information regarding syntactic dependencies. Word $i$ at layer $l$ can be represented as ${\bf h}^l_i$. This graphical structure is depicted as
\begin{equation}
\label{eq:gcn1}
{\bf h}^l_i = \sigma (\sum_{j=1}^{n}\tilde{\mathbf{A}}_{ij}\mathbf{W}^{l}h_{j}^{l-1}+\mathbf{b}^{l}).
\end{equation}
In GCN, $\mathbf{W}^l$ is the weight for linear transformation, $\mathbf{b}^l$ acts as bias, and $\sigma$ is the nonlinear activation function. GCN utilizes the sentence's dependency tree to enforce syntactic constraints, facilitating the identification of descriptive words associated with a specific aspect, considering their syntactic proximity. This capability enables GCN to effectively handle non-adjacent words related to aspect sentiment, making it a suitable choice for aspect-based sentiment analysis. The update of each word's representation is executed through Bi-GCN as follows:
\begin{equation}
\label{eq:gcn2}
\overrightarrow {\bf h}^l_i =\sum_{j=1}^{n}\tilde{\mathbf{A}}_{ij}\mathbf{W}^{l}h_{j}^{l-1}~~\text{and}~~
\overleftarrow {\bf h}^l_i = \sum_{j=1}^{n}{\tilde{\mathbf{A}}_{ij}}^{T} \mathbf{W}^{l}h_{j}^{l-1},
\end{equation}
where $\overrightarrow {\bf h}^l_i$ and $\overleftarrow {\bf h}^l_i$ are the outputs of the forward and backward hidden layers for the word $i$ in the $l$th layer. The forward and backward representations are concatenated by
\begin{equation}
\label{eq:gcn4}
\tilde{\bf h}^l_i = \overrightarrow {\bf h}^l_i \oplus \overleftarrow {\bf h}^l_i
~~\text{and}~~
{\bf h}^l_i = \text{ReLU}(\frac{\tilde{\bf h}^l_i}{d_i+1}\mathbf{W}^{l}+\mathbf{b}^{l}),
\end{equation}
where $d_i$ represents the degree of the $i$th token in the adjacency matrix. Note that the parameters $\mathbf{W}^{l}$ and $\mathbf{b}^{l}$ are trainable in the Bi-GCN network.

\subsection{Aspect-Specific Masking}
\label{sec:masking}
We utilize the masking technique to reduce redundancy in the Bi-GCN hidden representation by hiding aspect-independent text. This masking can be depicted as ${\bf h}_i^l = 0,~1\leq i< \gamma,~\gamma +m-1<t\leq n$. The result from the zero mask layer consists of aspect-focused characteristics, represented as $\mathbf{H}_{mask}^l = \left\{{\bf 0}, \cdots, {\bf h}_{\gamma}^l , \cdots, {\bf h}_{\gamma+m-1}^l, \cdots, {\bf 0} \right\}$. Utilizing graph convolution, $\mathbf{H}_{mask}^l$ adeptly captures contextual information, including syntactic dependencies and extended multi-word relationships. The attention mechanism for the hidden layer operates as follows:

\begin{equation}
\label{eq:mask3}
\beta_i=\sum_{j=1}^{n}{\bf h}_j^T {\bf h}_j^l~~\text{and}~~
\alpha_i=\frac{\exp(\beta_i)}{\sum_{j=1}^n\exp(\beta_j))}.
\end{equation}
Finally, the sentiment classification combines cumulative representation scores with attentions from the transformer, resulting in the sentiment classification:
\begin{equation}
\label{eq:mask5}
\mathbf{res}_{out} = \sum_{i=1}^n\alpha_i{\bf h}_i + {\bf Z}_{out}.
\end{equation}

\subsection{Sentiment Classification}
\label{sec:classification}
The computed $\mathbf{r}{out}$ from the previous step serves as input to the fully connected layer for sentiment classification, given by $\mathbf{prob}= \text{softmax}(\mathbf{r}{out}\mathbf{W}_p + \mathbf{b})$. Finally, utilizing cross-entropy and $\mathit{L_{2}}$ regularization \cite{cortes20092}, the loss function is as follows:
\begin{equation}
\label{eq:train}
\text{Loss} = - \sum_{\hat{\mathit{p}}\in C}\log\mathbf{prob_{\hat{\mathit{p}}}}+ \lambda \left \| \Theta  \right \|_{2},
\end{equation}
where $\mathit{p}$ signifies the label, and $C$ represents the dataset. $\mathbf{prob_{\hat{\mathit{p}}}}$ denotes the $\hat{\mathit{p}}$th element of $\mathbf{prob}$. $\Theta$ and $\lambda$ are the trainable parameters.

\section{Experiments}
\label{sec:exp}
\subsection{Datasets}
\label{sec:datasets}
\begin{table}[ht]
\renewcommand\arraystretch{1.2}
\renewcommand\tabcolsep{12pt} 
\caption{Statistics of the four datasets.}
\label{tab:statistics}
\centering
\begin{tabular}{ccccc}
\toprule
Dataset & Split & Positive & Neutral & Negative \\\midrule
\multirow{2}[4]{*}{Twitter} & Train & 1561  & 3127  & 1560 \\   
& Test  & 173   & 346   & 173 \\\midrule
\multirow{2}[4]{*}{Rest14} 
& Train & 2164 & 637 & 807 \\         
& Test  & 728   & 196   & 196 \\\midrule
\multirow{2}[4]{*}{Rest15} & Train & 912   & 36    & 256 \\
& Test  & 326   & 34    & 182 \\\midrule
\multirow{2}[4]{*}{Rest16} & Train & 1240  & 69    & 439 \\
& Test  & 469   & 30    & 117 \\\bottomrule
\end{tabular}
\end{table}
Four commonly used benchmark datasets were employed to assess the performance of the SentiSys model. The Twitter dataset \cite{dong2014adaptive} encompasses 6,940 Twitter posts from users, with 6,248 data points for training and 692 for testing. Additionally, three datasets, namely Rest14 \cite{pontiki-etal-2014-semeval}, Rest15 \cite{pontiki2015semeval}, and Rest16 \cite{pontiki2016semeval} contain customer comments and restaurant rating information. Rest14 comprises 4,728 customer comments, with 3,608 for training and 1,120 for testing. Rest15 includes 1,746 reviews, with 1,204 for training and 542 for testing. Rest16, the 2016 benchmark dataset, consists of 2,364 reviews, of which 1,748 are used for training, and the remaining 616 are for testing. To enhance dataset quality, sentences with conflicting polarity or unclear meaning were removed from the Rest15 and Rest16 datasets. Table \ref{tab:statistics} shows the statistics of the four datasets.

\subsection{Experimental Setup}
\label{sec:setting}
We initialize the word embedding layer using a pre-trained GloVe vectors \cite{pennington2014glove} with a dimensionality of 300. The hidden layer of the Bi-LSTM also has a dimensionality of 300. The Bi-GCN of SentiSys contains 3 layers with parameters initialized using a uniform distribution. The optimizer is Adam \cite{zhang2018improved}, which has a learning rate of 0.001. We use $\mathit{L_{2}}$ regularization with a batch size of 32 for up to 100 epochs.

\begin{table}[t]
\caption{Results of the comparison of the performance of baseline models with that of SentiSys. The values in gray cells indicate the maximum values.}
\label{tab:result}
\centering
\renewcommand\arraystretch{1.2}
\resizebox{\textwidth}{35mm}{
\begin{tabular}{@{}lllllllll@{}}\toprule
\multirow{2}{*}{Model} & \multicolumn{2}{l}{Twitter} & \multicolumn{2}{l}{Rest14} & \multicolumn{2}{l}{Rest15} & \multicolumn{2}{l}{Rest16} \\ \cmidrule(l){2-9} 
&Acc (\%) $\uparrow$ &F1 (\%) $\uparrow$&Acc (\%) $\uparrow$&F1 (\%) $\uparrow$&Acc (\%) $\uparrow$&F1 (\%) $\uparrow$&Acc (\%) $\uparrow$&F1 (\%) $\uparrow$\\ \cmidrule(r){1-9}
SVM&63.40&63.30&80.16&-&-&-&-&-\\
LSTM&69.56&67.70&78.13&67.47&77.37&55.17&86.80&63.88\\
TD-LSTM&70.80&69.00&78.00&66.73&76.39&58.70&82.16&54.21\\
MemNet&71.48&69.90&79.61&69.64&77.31&58.28&85.44&65.99\\
AOA&72.30&70.20&79.97&70.42&78.17&57.02&87.50&66.21\\
IAN&72.50&70.81&79.26&70.09&78.54&52.65&84.74&55.21\\
TNet-LF&\colorbox[gray]{0.9}{72.98}&\colorbox[gray]{0.9}{71.43}&80.42&71.03&78.47&59.47&\colorbox[gray]{0.9}{89.07}&\colorbox[gray]{0.9}{70.43}\\
ASGCN&71.53&69.68&80.86&72.19&79.34&60.78&88.69&66.64\\ 
MGAN&72.54&70.81&81.25&71.94&79.36&57.26&87.06&62.29\\
Sentic LSTM&70.66&67.87&79.43&70.32&79.55&60.56&83.01&68.22\\
CMLA-ALSTM&-&-&77.46&63.87&81.03&54.79&-&-\\
RACL&-&-&81.42&69.59&\colorbox[gray]{0.9}{83.26}&59.85&-&-\\\midrule
SentiSys &72.44&70.67&\colorbox[gray]{0.9}{81.70}&\colorbox[gray]{0.9}{73.63}&79.34&\colorbox[gray]{0.9}{61.99}&88.80&68.96\\\toprule   
\end{tabular}}
\end{table}

\subsection{Evaluation Measures}
\label{sec:measures}
We use two performance metrics, Accuracy (Acc) and Macro F1 (F1), to evaluate and validate the effectiveness of the proposed SentiSys. Acc and F1 are widely recognized in the field of multi-classification tasks to assess the predictive and classification capabilities of models \cite{zhang2020hierarchy}. Concretely, Acc quantifies the ratio of correctly classified samples and thus directly measures the accuracy of the model. F1 takes into account both accuracy and recall, offering a comprehensive perspective on the model's performance.

\subsection{Evaluation Results}
\label{sec:results}
Table \ref{tab:result} presents a comparative analysis of the Acc and F1 scores between the proposed SentiSys model and 15 other baseline models. Notably, the SentiSys model demonstrates superior performance across various metrics in the Twitter, Rest14, and Rest15 datasets. 
While the TNet-LF model achieved the highest Acc and F1 scores in the Twitter and Rest16 datasets, it did not exhibit the same level of effectiveness as the SentiSys model in the Rest14 and Rest15 datasets. Specifically, on the Rest14 dataset, the SentiSys model stands out, achieving significantly higher Acc and F1 scores compared to the other baseline models. In the Rest15 dataset, RACL achieved the highest Acc score, but its F1 score was relatively low at 59.85. When it comes to the Rest16 dataset, the Acc and F1 scores of TNet-LF were only marginally superior to those of the SentiSys model. This observation could be attributed to the critical role of information from both parent and child nodes, highlighting the significance of considering the entire graph structure rather than solely processing it as a directed graph. Additionally, the limited sentence grammar in the Twitter dataset may have constrained prediction results. In summary, the experimental results show that our proposed SentiSys enhances the performance of aspect-based sentiment analysis systems compared with other baseline models, especially for TNet-LF on the Rest16 dataset. Furthermore, SentiSys outperforms ASGCN on all four benchmark datasets, which highlights the impact of word dependence on enhancing the classification results.


\subsection{Effect of GCN Layers}
\label{sec:layers}

\begin{figure}[t]  
\centering  
\includegraphics[width=0.7\hsize]{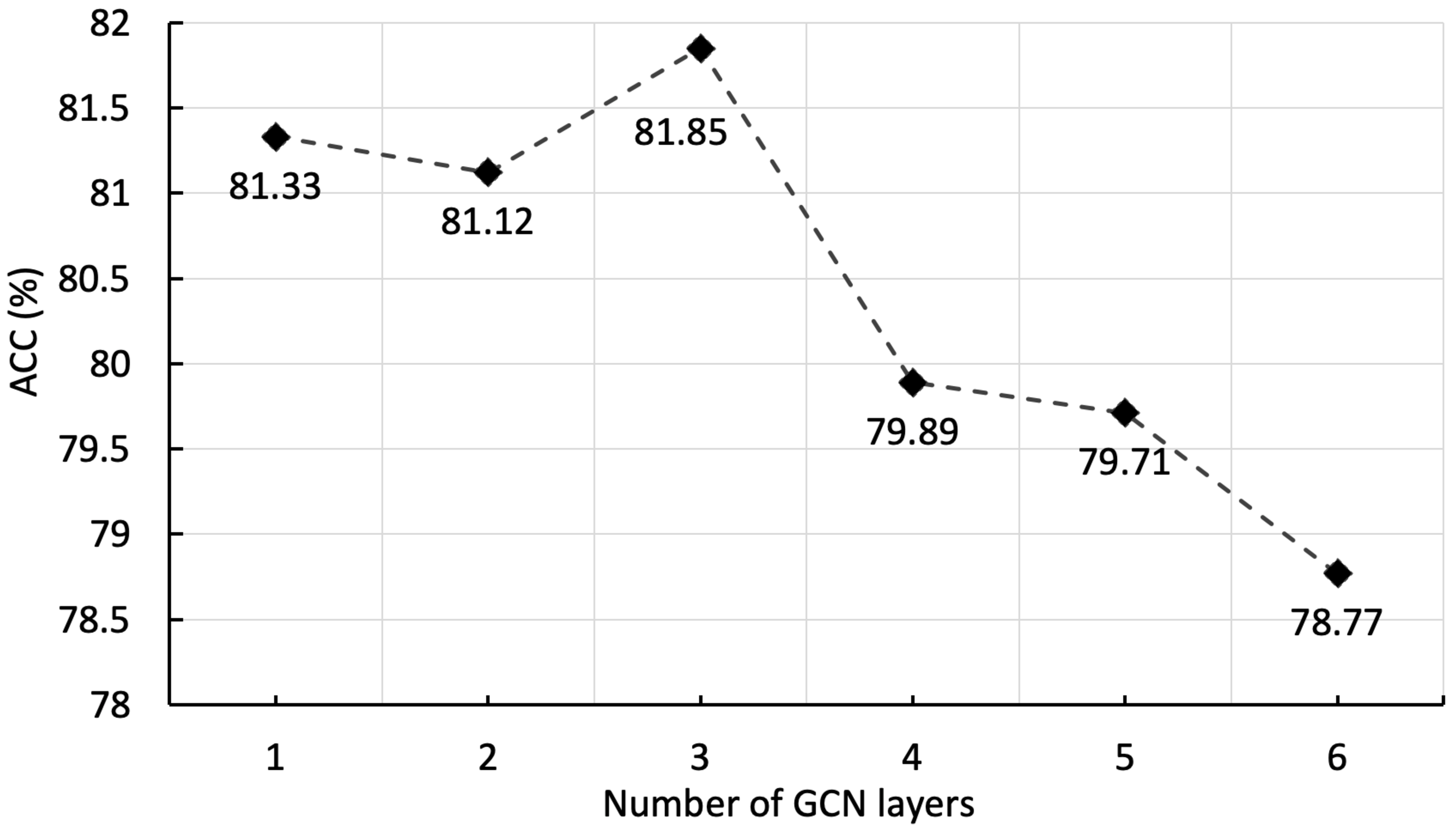}   
\caption{Change curve of Acc scores with the number of GCN layers.}   
\label{fig:acc} 

\includegraphics[width=0.7\hsize]{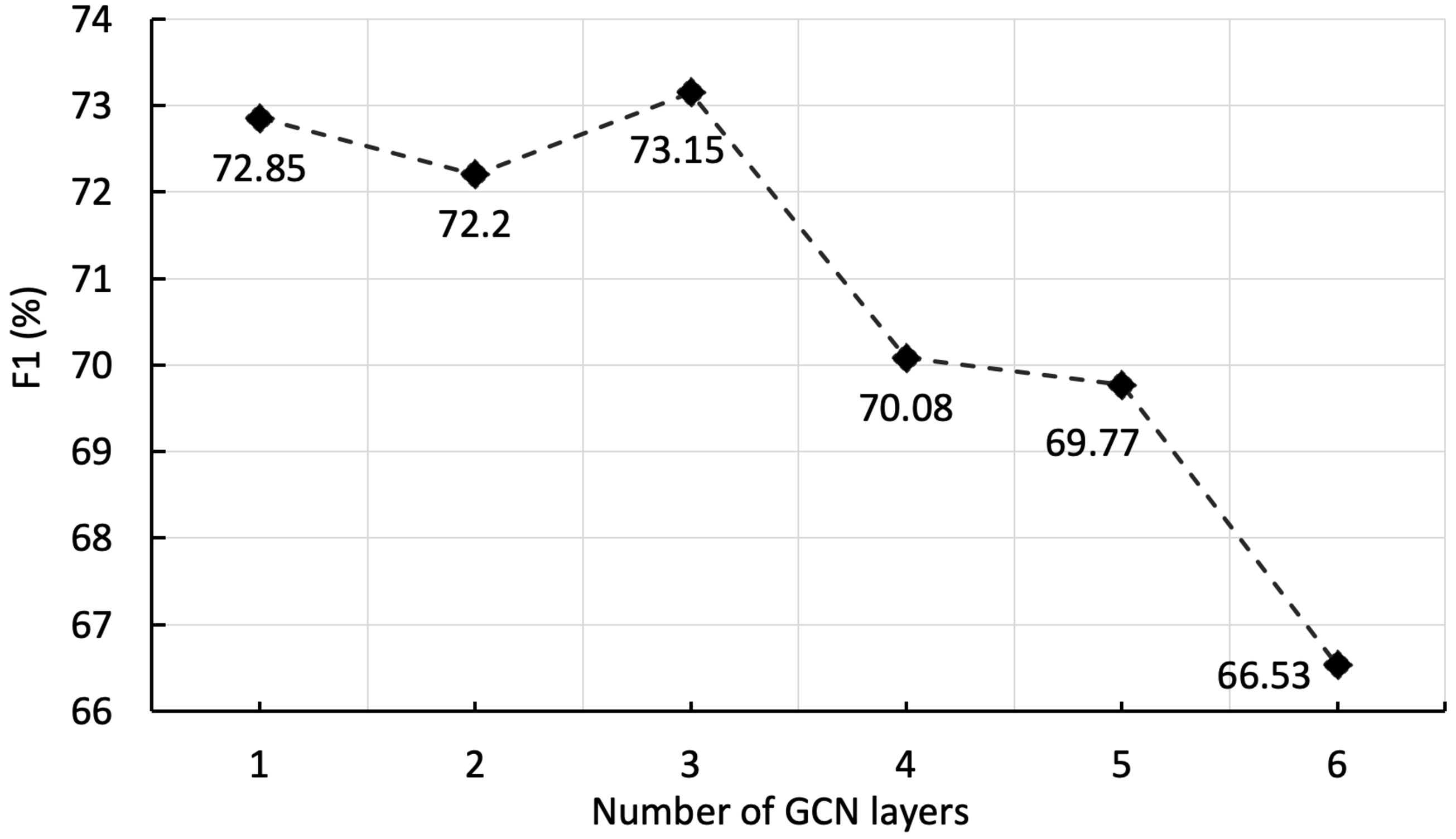}   
\caption{Change curve of F1 scores with the number of GCN layers.}   
\label{fig:f1} 
\end{figure}

Figures \ref{fig:acc} and \ref{fig:f1} show the effects of different GCN layers on the Acc and F1 scores of the proposed SentiSys model on the Rest14 dataset. X and Y axes indicate the number of GCN layers and the corresponding Acc and F1 scores, respectively. The experimental results show that the Acc and F1 scores tended to increase and then decrease as the number of layers increased. The Acc and F1 scores achieved the maximum values with the number of GCN layers of 3. The scores are 0.818 and 0.731, respectively.

\section{Conclusion}
\label{sec:conclusion}
This study improved aspect-based sentiment analysis with SentiSys, using an edge-enhanced bidirectional GCN. Bi-LSTM provided deep text understanding, with hidden features used to construct syntactic dependency trees in Bi-GCN. SentiSys included a transformer encoder layer for global information extraction, enhancing feature capturing in lengthy text. The dependency parsing layer within Bi-GCN extracted graph structure based on text syntax, aiding aspect-context connections learning. An aspect-specific masking layer refined the model, reducing redundant information in GCN's hidden layer and enhancing accuracy. Finally, SentiSys combined Bi-GCN and transformer outputs for sentiment classification. Two primary limitations were identified: statistical computation of SDI and transformer's feature enhancement stage. Future work aims to address these limitations by integrating trainable vectors for dynamic SDI computation and exploring optimal strategies for transformer integration to enhance classification accuracy.

\section*{Acknowledgements}
This research was supported by the National Natural Science Foundation of China (Grant No.  62402022), the International Research Fellow of Japan Society for the Promotion of Science (Postdoctoral Fellowships for Research in Japan [Standard]), and KAKENHI (20K11830) Japan.

\bibliographystyle{splncs04}
\bibliography{refs}

\begin{thebibliography}{10}
\providecommand{\url}[1]{\texttt{#1}}
\providecommand{\urlprefix}{URL }
\providecommand{\doi}[1]{https://doi.org/#1}

\bibitem{augustyniak2021comprehensive}
Augustyniak, {\L}., Kajdanowicz, T., Kazienko, P.: Comprehensive analysis of
  aspect term extraction methods using various text embeddings. Computer Speech
  \& Language  \textbf{69},  101217 (2021)

\bibitem{bauman2022know}
Bauman, K., Tuzhilin, A.: Know thy context: Parsing contextual information from
  user reviews for recommendation purposes. Information Systems Research
  \textbf{33}(1),  179--202 (2022)

\bibitem{cai2020aspect}
Cai, H., Tu, Y., Zhou, X., Yu, J., Xia, R.: Aspect-category based sentiment
  analysis with hierarchical graph convolutional network. In: Proceedings of
  the 28th International Conference on Computational Linguistics. pp. 833--843
  (2020)

\bibitem{chang2023reducing}
Chang, M., Yang, M., Jiang, Q., Xu, R.: Reducing spurious correlations for
  aspect-based sentiment analysis with variational information bottleneck and
  contrastive learning. ArXiv Preprint ArXiv:2303.02846  (2023)

\bibitem{chen2017recurrent}
Chen, P., Sun, Z., Bing, L., Yang, W.: Recurrent attention network on memory
  for aspect sentiment analysis. In: Proceedings of the 2017 Conference on
  Empirical Methods in Natural Language Processing. pp. 452--461 (2017)

\bibitem{chen2020relation}
Chen, Z., Qian, T.: Relation-aware collaborative learning for unified
  aspect-based sentiment analysis. In: Proceedings of the 58th Annual Meeting
  of the Association for Computational Linguistics. pp. 3685--3694 (2020)

\bibitem{cortes20092}
Cortes, C., Mohri, M., Rostamizadeh, A.: L2 regularization for learning
  kernels. In: Proceedings of the Twenty-Fifth Conference on Uncertainty in
  Artificial Intelligence. pp. 109--116 (2009)

\bibitem{ding2008holistic}
Ding, X., Liu, B., Yu, P.S.: A holistic lexicon-based approach to opinion
  mining. In: Proceedings of the 2008 International Conference on Web Search
  and Data Mining. pp. 231--240 (2008)

\bibitem{do2019deep}
Do, H.H., Prasad, P.W., Maag, A., Alsadoon, A.: Deep learning for aspect-based
  sentiment analysis: A comparative review. Expert systems with applications
  \textbf{118},  272--299 (2019)

\bibitem{dong2014adaptive}
Dong, L., Wei, F., Tan, C., Tang, D., Zhou, M., Xu, K.: Adaptive recursive
  neural network for target-dependent twitter sentiment classification. In:
  Proceedings of the 52nd Annual Meeting of the Association for Computational
  Linguistics. pp. 49--54 (2014)

\bibitem{graves2013speech}
Graves, A., Mohamed, A.r., Hinton, G.: Speech recognition with deep recurrent
  neural networks. In: 2013 IEEE International Conference on Acoustics, Speech
  and Signal Processing. pp. 6645--6649 (2013)

\bibitem{hu2021graph}
Hu, Y., Shen, H., Liu, W., Min, F., Qiao, X., Jin, K.: A graph convolutional
  network with multiple dependency representations for relation extraction.
  IEEE Access  \textbf{9},  81575--81587 (2021)

\bibitem{jiang2011target}
Jiang, L., Yu, M., Zhou, M., Liu, X., Zhao, T.: Target-dependent twitter
  sentiment classification. In: Proceedings of the 49th Annual Meeting of the
  Association for Computational Linguistics: Human Language Technologies. pp.
  151--160 (2011)

\bibitem{kipf2016semi}
Kipf, T.N., Welling, M.: Semi-supervised classification with graph
  convolutional networks. ArXiv Preprint ArXiv:1609.02907  (2016)

\bibitem{laskari2016aspect}
Laskari, N.K., Sanampudi, S.K.: Aspect based sentiment analysis survey. IOSR
  Journal of Computer Engineering (IOSR-JCE)  \textbf{18}(2),  24--28 (2016)

\bibitem{li2023spotgan}
Li, C., Yamanishi, Y.: {SpotGAN}: A reverse-transformer gan generates
  scaffold-constrained molecules with property optimization. In: Joint European
  Conference on Machine Learning and Knowledge Discovery in Databases. pp.
  323--338. Springer (2023)

\bibitem{li2024gxvaes}
Li, C., Yamanishi, Y.: {GxVAEs:} two joint vaes generate hit molecules from
  gene expression profiles. In: Proceedings of the AAAI Conference on
  Artificial Intelligence. vol.~38, pp. 13455--13463 (2024)

\bibitem{li2024tengan}
Li, C., Yamanishi, Y.: {TenGAN}: Pure transformer encoders make an efficient
  discrete gan for de novo molecular generation. In: International Conference
  on Artificial Intelligence and Statistics. pp. 361--369. PMLR (2024)

\bibitem{li2018transformation}
Li, X., Bing, L., Lam, W., Shi, B.: Transformation networks for target-oriented
  sentiment classification. In: Proceedings of the 56th Annual Meeting of the
  Association for Computational Linguistics. pp. 946--956 (2018)

\bibitem{liao2021improved}
Liao, W., Zeng, B., Yin, X., Wei, P.: An improved aspect-category sentiment
  analysis model for text sentiment analysis based on {RoBERTa}. Applied
  Intelligence  \textbf{51},  3522--3533 (2021)

\bibitem{ma2018targeted}
Ma, Y., Peng, H., Cambria, E.: Targeted aspect-based sentiment analysis via
  embedding commonsense knowledge into an attentive {LSTM}. In: Proceedings of
  the AAAI Conference on Artificial Intelligence. vol.~32 (2018)

\bibitem{meng2020structure}
Meng, F., Feng, J., Yin, D., Chen, S., Hu, M.: A structure-enhanced graph
  convolutional network for sentiment analysis. In: Findings of the Association
  for Computational Linguistics: EMNLP 2020. pp. 586--595 (2020)

\bibitem{pennington2014glove}
Pennington, J., Socher, R., Manning, C.D.: Glove: Global vectors for word
  representation. In: Proceedings of the 2014 Conference on Empirical Methods
  in Natural Language Processing (EMNLP). pp. 1532--1543 (2014)

\bibitem{phan2020modelling}
Phan, M.H., Ogunbona, P.O.: Modelling context and syntactical features for
  aspect-based sentiment analysis. In: Proceedings of the 58th annual meeting
  of the association for computational linguistics. pp. 3211--3220 (2020)

\bibitem{pontiki2015semeval}
Pontiki, M., Galanis, D., Papageorgiou, H., Manandhar, S., Androutsopoulos, I.:
  Semeval-2015 task 12: Aspect based sentiment analysis. In: Proceedings of the
  9th International Workshop on Semantic Evaluation (SemEval 2015). pp.
  486--495 (2015)

\bibitem{pontiki2016semeval}
Pontiki, M., Galanis, D., Papageorgiou, H., Androutsopoulos, I., Manandhar, S.,
  AL-Smadi, M., Al-Ayyoub, M., Zhao, Y., Qin, B., De~Clercq, O., et~al.:
  Semeval-2016 task 5: Aspect based sentiment analysis. In: Proceedings of the
  Workshop on Semantic Evaluation (SemEval-2016). pp. 19--30 (2016)

\bibitem{pontiki-etal-2014-semeval}
Pontiki, M., Galanis, D., Pavlopoulos, J., Papageorgiou, H., Androutsopoulos,
  I., Manandhar, S.: {S}em{E}val-2014 task 4: Aspect based sentiment analysis.
  In: Proceedings of the 8th International Workshop on Semantic Evaluation
  ({S}em{E}val 2014). pp. 27--35 (2014)

\bibitem{rani2022efficient}
Rani, S., Bashir, A.K., Alhudhaif, A., Koundal, D., Gunduz, E.S., et~al.: An
  efficient {CNN-LSTM} model for sentiment detection in \#blacklivesmatter.
  Expert Systems with Applications  \textbf{193},  116256 (2022)

\bibitem{tang2016effective}
Tang, D., Qin, B., Feng, X., Liu, T.: Effective {LSTM}s for target-dependent
  sentiment classification. In: Proceedings of the 26th International
  Conference on Computational Linguistics: Technical Papers. pp. 3298--3307
  (2016)

\bibitem{vaswani2017attention}
Vaswani, A., Shazeer, N., Parmar, N., Uszkoreit, J., Jones, L., Gomez, A.N.,
  Kaiser, {\L}., Polosukhin, I.: Attention is all you need. Advances in Neural
  Information Processing Systems  \textbf{30} (2017)

\bibitem{wang2020relational}
Wang, K., Shen, W., Yang, Y., Quan, X., Wang, R.: Relational graph attention
  network for aspect-based sentiment analysis. In: Proceedings of the 58th
  Annual Meeting of the Association for Computational Linguistics. pp.
  3229--3238 (2020)

\bibitem{wang2016attention}
Wang, Y., Huang, M., Zhu, X., Zhao, L.: Attention-based {LSTM} for aspect-level
  sentiment classification. In: Proceedings of the 2016 Conference on Empirical
  Methods in Natural Language Processing. pp. 606--615 (2016)

\bibitem{zhang2019aspect}
Zhang, C., Li, Q., Song, D.: Aspect-based sentiment classification with
  aspect-specific graph convolutional networks. In: Proceedings of the 2019
  Conference on Empirical Methods in Natural Language Processing and the 9th
  International Joint Conference on Natural Language Processing (EMNLP-IJCNLP).
  pp. 4568--4578 (2019)

\bibitem{zhang2019predicting}
Zhang, J., Hu, X., Jiang, Z., Song, B., Quan, W., Chen, Z.: Predicting
  disease-related {RNA} associations based on graph convolutional attention
  network. In: 2019 IEEE International Conference on Bioinformatics and
  Biomedicine (BIBM). pp. 177--182. IEEE (2019)

\bibitem{zhang2020hierarchy}
Zhang, J., Jiang, Z., Du, Y., Li, T., Wang, Y., Hu, X.: Hierarchy construction
  and classification of heterogeneous information networks based on rsdaef.
  Data \& Knowledge Engineering  \textbf{127},  101790 (2020)

\bibitem{zhang2024cd}
Zhang, J., Wang, S., Jiang, Z., Chen, Z., Bai, X.: {CD-Net}: Cascaded 3d
  dilated convolutional neural network for pneumonia lesion segmentation.
  Computers in Biology and Medicine p. 108311 (2024)

\bibitem{zhang2024quantitative}
Zhang, J., Wang, Z., Jiang, Z., Wu, M., Li, C., Yamanishi, Y.: Quantitative
  evaluation of molecular generation performance of graph-based gans. Software
  Quality Journal pp. 1--29 (2024)

\bibitem{zhang2019multi}
Zhang, X., Li, C., Morimoto, Y.: A multi-factor approach for stock price
  prediction by using recurrent neural networks. Bulletin of networking,
  computing, systems, and software  \textbf{8}(1),  9--13 (2019)

\bibitem{zhang2018improved}
Zhang, Z.: Improved {A}dam optimizer for deep neural networks. In: 2018
  IEEE/ACM 26th International Symposium on Quality of Service (IWQoS). pp.~1--2
  (2018)

\bibitem{zhou2020graph}
Zhou, J., Cui, G., Hu, S., Zhang, Z., Yang, C., Liu, Z., Wang, L., Li, C., Sun,
  M.: Graph neural networks: A review of methods and applications. AI open
  \textbf{1},  57--81 (2020)

\end{thebibliography}
\end{document}